# ScamSpot: Fighting Financial Fraud in Instagram Comments

Stefan Erben[1,2], Andreas Waldis[1,3]

[1]Lucerne University of Applied Sciences and Arts
[2]University of Applied Sciences Technikum Vienna
[3]Ubiquitous Knowledge Processing Lab (UKP Lab), Department of Computer Science and Hessian Center for AI (hessian.AI), Technical University of Darmstadt

`<office@stefanerben.com>`

## Abstract

The long-standing problem of spam and fraudulent messages in the comment sections of Instagram pages in the financial sector claims new victims every day. Instagram's current spam filter proves inadequate, and existing research approaches are primarily confined to theoretical concepts. Practical implementations with evaluated results are missing. To solve this problem, we propose ScamSpot, a comprehensive system that includes a browser extension, a fine-tuned BERT model and a REST API. This approach ensures public accessibility of our results for Instagram users using the Chrome browser. Furthermore, we conduct a data annotation study, shedding light on the reasons and causes of the problem and evaluate the system through user feedback and comparison with existing models. ScamSpot is an open-source project and is publicly available at https://scamspot.github.io/.

## 1 Introduction

Financial fraud has switched its medium – from phone calls and emails to social media (Ramli et al., 2023; Soomro & Hussain, 2019). A recent report from the U.S. Federal Trade Commission shows that the number of social media scams has soared in recent years and especially cryptocurrency scams are initiated on Instagram[1].

Last year, the topic also gained political attendance[2], but so far, Instagram users have not seen any improvements (Table 8) and industry experts continue to voice their concerns about the spam and scam problem (Kerr et al., 2023). While spam is defined as unsolicited or unwanted content/comments (Hayati et al., 2010), scam is characterized as deceptive or fraudulent activity resulting in mostly financial loss for the victim (Liebau & Schueffel, 2019). New and inexperienced investors in particular fall victim to targeted attacks, often losing significant sums of money in the process[1].

Instagram's existing spam filter has a precision of 98.36%, but only a recall of 11.51% (Table 6). Existing research mentioned in Section 6 made its first success in theoretical concepts and general spam detection, yet no practical solutions to detect financial spam and scam comments have been published. The problem has not been solved, as 90% of our survey participants expressed their dissatisfaction (Table 8), showcasing the urgency for a solution.

To close this gap, we show a way to efficiently classify comments with high precision and communicate the results to the user in real-time to improve the user experience and reduce the likelihood of fraud incidents. The solution is ScamSpot, a system designed to remove spam and scam comments from Instagram. More specifically, we contribute as follows:

- **Dataset & Data Annotation**: We compile what we believe to be the first large dataset of over 100,000 comments focusing specifically on Instagram comments of the financial space (Sections 2 and 3). We annotate over 3,000 comments as part of a data annotation study, which shows that domain-specific knowledge is needed to accurately classify the comments. By making all our data publicly available, we enable further research in this area.

---

[1] https://www.ftc.gov/news-events/data-visualizations/data-spotlight/2022/06/reports-show-scammers-cashing-crypto-craze

[2] https://www.feinstein.senate.gov/public/index.cfm/press-releases?ID=C51E17BC-D39D-4913-AA6C-09AB6B259083

- **ScamSpot System**: Our core solution, encompasses a fine-tuned BERT model (Section 4.1), a user-friendly Chrome browser extension (Section 4.2) and a REST API (Section 4.3). This enables the detection and removal of fraudulent comments on Instagram in real-time.

- **Systematic Evaluation**: We evaluate ScamSpot in two cycles, both quantitatively and qualitatively (Section 5). Our results demonstrate dramatic improvements in usability and increased user satisfaction, emphasising the relevancy of ScamSpot.

The result is an evaluated and deployed application that enables every Instagram user to use the website with a significantly reduced risk of encountering spam and fraud, making the user experience more enjoyable and secure.

## 2 Dataset & Data Annotation Study

The development of ScamSpot necessitates the creation of a specialized dataset, as existing research lacks robust data samples for Instagram comments, particularly in the financial sector. Recognizing the gap, we embark on a two-fold mission: not only to gather this essential data but also to annotate it meticulously, thereby contributing a valuable resource to the community.

To address this need, we develop a Python script utilizing an existing library[3] to access Instagram's private API. Between February 28th and May 4th 2023, we collected data from 38 Instagram pages related to finance and cryptocurrencies. This effort yields a dataset of over 100,000 comments, which, to our knowledge, represents one of the largest publicly available datasets in this niche. We have made this dataset, along with the scraping script, openly accessible, enabling others to benefit from and expand upon our work. Comment examples can be found in Table 7 in the appendix.

The pivotal aspect of our study was the annotation of 3,445 comments (66.6% genuine, 33.4% spam/scam). We annotate the dataset ourselves using a simple, self-developed web interface. While one of the team members has several years of experience as an owner of multiple large financial Instagram pages, the quality of the annotated comments was validated by continuously checking a subset of comments against the later hand-picked experts' classifications.

To better understand how expertise in the financial sector influences the identification of spam and scams we perform a data annotation study. We divide participants into two groups: 'experts' with substantial industry knowledge and 'amateurs' with less or no such experience. This distinction is critical, as it highlights the challenges faced by non-experts in recognizing fraudulent content, a key factor in the importance of ScamSpot.

While experts reach a Fleiss Kappa agreement of 0.618, amateurs manage only 0.519 (Fleiss, 1971). This disparity underscores the necessity of expert knowledge in accurately classifying such comments. To solidify these results, we conduct a follow-up study with 11 handpicked industry experts, resulting in a near-perfect Fleiss Kappa score of 0.808. These results not only validate our approach but also emphasize the nuanced differences in definitions of spam and scam among professionals.

The insights from this annotation study are instrumental in shaping ScamSpot. They not only inform the training of our models but also highlight the real-world challenge faced by everyday Instagram users, particularly the less experienced ones, in navigating financial fraud. This aspect is later also mirrored in the performance of advanced language models like GPT-3 and GPT-4, which also struggle with categorizing these comments, further accentuating the complexity of the task and the value of our expert-driven approach.

To sum up, our efforts in data collection and annotation are not just preliminary steps but foundational to the development of ScamSpot. By making these resources publicly available, we aim to facilitate further research in this vital area, emphasizing the importance of domain-specific expertise in combating financial fraud on social media platforms.

---

[3] https://github.com/adw0rd/instagrapi

## 3  Model Considerations

The next objective is to demonstrate an effective system rather than conduct an exhaustive analysis of various models. Nevertheless, we test a variety of models to identify an effective solution for detecting fraudulent Instagram comments. The selection of models includes traditional statistical models, large language models and BERT as a representative of the transformer models. The following specific models are selected:

- **Statistical Models**: We start with a linear regression, a decision tree and a random forest model which provide us with a baseline and show that they lack the sophistication needed for our complex requirements.

- **Large Language Models (LLMs)**: We assess the advanced natural language processing capabilities of GPT-3 ("`gpt-3.5-turbo`") and GPT-4 ("`gpt-4-1106-preview`"). Their ability to understand and generate human-like text is a key consideration in our analysis. However, the lack of transparency and control over the training data of these models is a major limitation. Despite the adjustment of seed and temperature parameters (Table 5), the results are not deterministic. Three test runs are made for each model and the results can be found in Table 5. The best results are documented in Table 1.

- **Transformer Models**: As a representative, we select BERT ("`bert-base-cased`") and fine-tune it based on the annotated data mentioned in Section 2. Its capability to understand context makes it a strong candidate in our selection.

In this study, we adopt a zero-shot approach with LLMs like GPT-3 and GPT-4, contrasting them with the fine-tuned BERT model. This methodology is chosen to demonstrate the practical usability of general-purpose LLMs in their standard configuration. While models like GPT-3 and GPT-4 show proficiency in general tasks, our findings align with those of Yu et al. (2023), when illustrating their limitations in specific tasks compared to fine-tuned models. This highlights the necessity of model selection tailored to task specificity and resource availability.

|       | Recall | Precision | F1     |
|-------|--------|-----------|--------|
| IG    | 0.1151 | **0.9836** | 0.2061 |
| DT    | 0.8032 | 0.7692    | 0.7859 |
| RF    | 0.8093 | 0.9244    | 0.8631 |
| LR    | 0.8353 | 0.9163    | 0.8740 |
| GPT-3 | 0.3739 | 0.3660    | 0.3699 |
| GPT-4 | 0.6348 | 0.5530    | 0.5911 |
| BERT  | **0.9213** | 0.9286 | **0.9249** |

Table 1: Model metrics. IG: Instagram's current spam filter, DT: Decision Tree, RF: Random Forest, LR: Linear Regression, BERT: Fined-tuned BERT model from ScamSpot

Despite the hype around large language models, the results in Table 1 show that even established models like a fine-tuned BERT model can drastically outperform newer models like GPT-4 or GPT-3. While both BERT and LLMs are transformer-based, our research also demonstrates that a fine-tuned BERT model is more successful in specific tasks, which is also shown by Yu et al. (2023).

This leads us to the decision to select a fine-tuned BERT model for ScamSpot. The reasons also include:

- **Promising Results**: Tests show that the fine-tuned BERT model far outperforms the other models in detecting fraudulent Instagram comments (Table 1).

- **Stability and Predictability**: Our fine-tuned BERT model demonstrates stable and predictable performance, a crucial factor for consistent user experience compared to the tested LLMs (Table 5).

- **Local Execution**: Unlike LLMs like GPT-3.5 and GPT-4, which often require external cloud services potentially exposing data to third parties, BERT can be deployed internally, ensuring data privacy. Moreover, BERT's smaller model size leads to lower energy consumption, an important consideration in sustainable AI development.

- **Adaptability to New Data**: BERT's ability to understand context and meanings beyond just keyword matching makes it superior to simpler models like TF-IDF vectors. This adaptability is vital for detecting variations in spam messages, ensuring our system remains effective in the face of evolving spam tactics.

In conclusion, our comprehensive testing and evaluation process leads us to choose a fine-tuned BERT model. The configuration for fine-tuning the model can be found in Table 3. This decision is informed by a balance of performance, stability, transparency, and futureproofing against evolving spam and scam strategies.

## 4 ScamSpot

To deliver comment classifications to users in real-time efficiently, we develop a Chrome browser extension, realizing the ScamSpot architecture. This approach is inspired by Rachmat (2018), which demonstrates the effectiveness of presenting classification results through a Chrome extension. Our approach differs by directly displaying the classifications without requiring users to do anything other than use the Instagram website. This approach allows the user to take full advantage of the AI model without compromising the user experience. Our system design consists of 3 main components.

The fine-tuned BERT model enables the classification of Instagram comments. The Chrome browser extension, which is installed by the user and runs on the user's device, allows us to extract the comments, communicate with our API, and manipulate the HTML DOM to display the results in real-time. A REST API acts as the connection between the browser extension and the BERT model.

### 4.1 BERT Model

One of the most important steps is to fine-tune a pre-trained BERT model for classifying Instagram comments (Devlin et al., 2019). BERT might not be the latest model, but the initial results are promising compared to other models (Table 1) and the process for implementation is straightforward. We also see that several other projects have had success in fine-tuning a BERT model in recent months with similar tasks (Sahmoud & Mikki, 2022; Santoso, 2022; Tida & Hsu, 2022). Figure 2 shows the model architecture and the associated specifications, Table 5 documents the fine-tuning of the BERT model including hyperparameters. We

---

[4] https://github.com/ScamSpot/scamspot_ml-models/

---

use the "`bert-base-cased`" model, which we train based on our annotated dataset of 3,445 comments (66.6% genuine, 33.4% spam or scam). Annotated comments are split 80:10:10 for training, validation, and testing. Both data as well as the code to fine-tune the model can be found as an open-source repository[4].

### 4.2 Chrome Browser Extension

To make the results accessible to the end user, we choose a Chrome browser extension. It is important for us to ensure a smooth and easy implementation for the user, which is possible with this approach. Through HTML scraping, comments are extracted locally from the user's browser and sent to our REST API. Based on the response, the comments

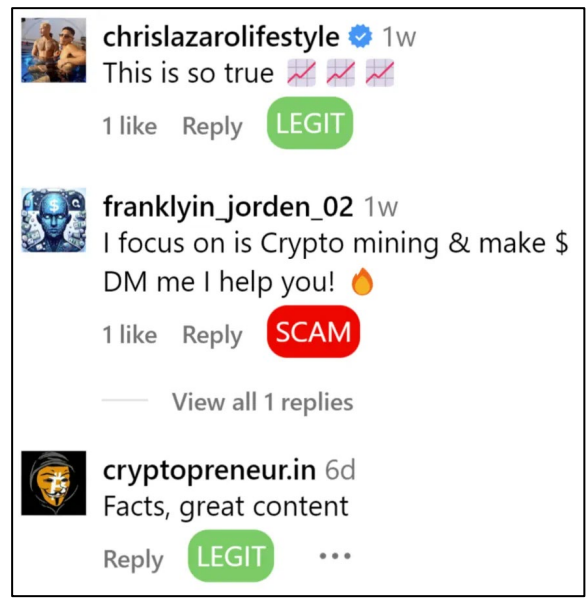

Figure 1: ScamSpot Mode 1, genuine and spam / scam comments are visually marked.

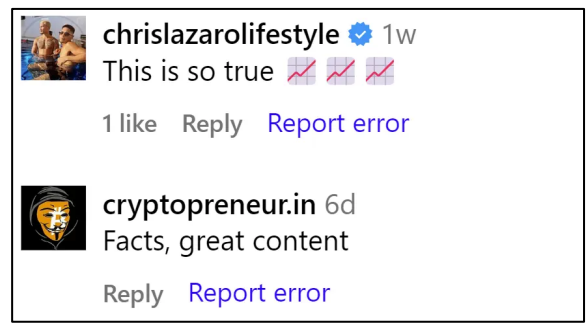

Figure 2: ScamSpot Mode 2, only genuine comments are visible, users can report invalid classifications.

in the HTML DOM are modified. The browser extension as well as its code is publicly available[5].

## 4.3 REST API

The REST API is developed using the Flask web framework and deployed utilizing Gunicorn, a Python Web Server Gateway Interface (WSGI) HTTP server. The server loads the fine-tuned BERT model and classifies the comment received on the `/scam` endpoint. The server's response indicates whether the comment is genuine or not. The code is available as a public repository[6]. Until now, the endpoints of the API have been accessible without user authentication/token access. This allows others to quickly use the API and prevents compliance issues. As usage increases, a user management system is planned, but as little data as possible should be stored.

## 4.4 Features

To use the application, the users need to install the Chrome browser extension and follow the brief installation guide[5]. After successful installation, the user can choose between two modes.

- **Detecting Fraudulent Comments**: The first mode of the application marks spam/fraud comments with a red label (Figure 1). Users are warned about a potential fraudulent comment, reducing the likelihood of a fraud case. However, the evaluation surveys conducted as part of the research show that users do not want to see spam or scam comments at all (Table 8).

- **Hiding Spam & Scam**: The second mode complies with these requests and hides all comments that are classified as spam or scam (Figure 2). The evaluation survey has clearly shown that users prefer the second mode, resulting in a drastically improved user experience.

- **Dynamic System**: Another valuable feature enables users to report incorrectly classified comments, contributing to model enhancement through subsequent data-driven refinement.

## 5 System Evaluation

ScamSpot's effectiveness is evaluated in two cycles based on Hevner's Design Science Research Framework (Hevner et al., 2004), which emphasizes the iterative development and refinement of a system through multiple cycles of design, testing, and feedback. In each cycle, both qualitative and quantitative evaluation metrics are considered.

The quantitative results focus on the metrics and the effectiveness of the model itself. As can be seen in Table 1, the model achieves both precision and recall of 92% after the second cycle.

Our model exhibits notable performance compared to other baseline models as well as OpenAI's GPT-3 and GPT-4, justifying its utilization in the project. The F1 scores can be found in Table 1, with the fine-tuned BERT model archiving a considerably higher F1 score.

When comparing the results with Instagram's existing spam filter, the model also performs well. While Instagram's spam filter only achieves a recall of 11.51%, our model achieves 92.13%. However, we must give credit to Instagram's spam filter as its precision was 98.36%, while our model only achieves 92.86%. This shows room for improvement.

Another notable result is that both GPT models fail to correctly categorise comments as spam/scam or genuine. Despite multiple approaches and different prompts, the results are surprisingly disappointing. Results can be found in Table 5.

Not only the quantitative but also the qualitative results improve drastically after the second cycle, in which the feedback from the first evaluation cycle was implemented. 20 ScamSpot users are asked in a questionnaire about their experiences with the system and their feedback.

When using the prototype, almost all test users report that the user experience has improved and that they are more willing to interact with others in the comments section. With the browser extension, users report overall positive feelings of joy and excitement compared to frustration and annoyance without ScamSpot, as shown in Figure 3.

---

[5] https://github.com/ScamSpot/scamspot_chrome-extension/

[6] https://github.com/ScamSpot/scamspot_api/

| | Method | Sample size | Prototype | Evaluation | F1 score |
|---|---|---|---|---|---|
| [1] (Septiandri & Wibisono, 2017) | SVM | 24,602 comments | × | × | 96.0% |
| [2] (Rachmat et al., 2018) | Prototype based on [1] | | ✓ | × | - |
| [3] (Haqimi et al., 2019) | CNB | 2,600 comments | × | × | 92.4% |
| [4] (Priyoko & Yaqin, 2019) | NB | 1,400 comments | × | × | 83.0% |
| [5] ScamSpot | BERT | 3,445 comments | ✓ | ✓ | 92.5% |

Table 2: Qualitative comparison of ScamSpot to previous works. SVM: Support Vector Machine, CNB: Complement Naive Bayes, NB: Naive Bayes

## 6 Related Work

In recent years, academic steps have already been taken regarding spam detection in social networks (Kaddoura et al., 2022). Most research has focused on Twitter, partly because it is easier to scrape data for training purposes and therefore ignored Instagram. Still, progress has also been made on Instagram, but more around account-based spam detection (Durga & Sudhakar, 2023; Kumar et al., 2023; Saranya Shree et al., 2021). The few comment-based approaches have so far focused on general spam, not financial fraud in Instagram comments.

A qualitative comparison of similar projects and papers can be found in Table 2. The first notable approach was in 2017, focusing on Indonesian comments with a large dataset and resulted in an F1 score of 96.0% (Septiandri & Wibisono, 2017). In 2019 two papers were published using smaller datasets while archiving an F1 score of 92.4% (Haqimi et al., 2019) and 83.0% (Priyoko & Yaqin, 2019).

The idea of using a browser extension and a REST API originated from a research team who based their work on the results of study [1] from Table 2. They constructed a prototype based on this approach (Rachmat et al., 2018).

Their system differentiates itself from ours since it focuses on comments under posts from Indonesian celebrities and users of the prototype still had to manually click on comments to check if they were spam. Our goal was to further improve the concept and ensure people see results directly within the website without having to click anything. Furthermore, a user evaluation also differentiates our work to ensure validated results.

The decision to fine-tune a BERT model (Devlin et al., 2019) was heavily influenced by the papers published shortly before our project which all archived great results in spam detection (Sahmoud & Mikki, 2022; Santoso, 2022; Tida & Hsu, 2022).

However, to our knowledge, there have been no studies looking at spam and fraud in comments under financial-related Instagram content.

## 7 Conclusion and Further Work

We introduce ScamSpot as an application that enables Instagram users to navigate the social media platform more safely by detecting fraudulent comments and removing unpleasant spam messages from the comment section of Instagram posts. We have shown that the combination of the fine-tuned BERT model and the Chrome browser extension results in a measurably better user experience and can reduce the number of fraud cases during active use.

Our solution is scalable and allows users with no technical background to safely use the application.

This work also contributes a scraped dataset of over 100,000 comments and an annotated dataset of 3,345 comments, which will help future projects and provide the results of our data annotation study. In addition, we evaluate our ScamSpot on both a quantitative and qualitative level and achieve excellent performance and a positive user impact.

To sum up, we show for the first time the viability and effectiveness of a fine-trained BERT for this classification problem, setting a precedent for future research on this issue. We further highlight the importance of combating spam and scams on Instagram, underscoring the need for solutions like ScamSpot.

In future developments, it is crucial to consider implementing more sophisticated machine learning models to enhance classification accuracy. This includes exploring the capabilities of newer variants of BERT and other encoder-only models such as DeBERTa, which have demonstrated superior performance in various classification tasks. Additionally, the potential of models like Mistral, Llama2, and their relatives should be investigated, especially in comparison to GPT models, to understand their effectiveness and efficiency in this context. Future research should also explore the effectiveness of LLMs by employing a non-zero-shot approach, customizing and training them specifically for this use case.

Moreover, further studies could focus on platforms other than Instagram, e.g., Twitter or YouTube, as their comments may be different but the concept is similar. Reducing the computational resources needed and thus reducing the cost and environmental impact is another approach for the future.

## Ethical Statement and Limitations

ScamSpot aims to promote a more equitable and inclusive financial system as well as protect new investors from fraudsters. We believe that the choice to take responsibility for one's financial future by starting to invest is a big step and that inexperienced investors should not be afraid of being scammed while they are still inexperienced and maybe a little naive.

All data generated and used in this study is publicly available and used under strict ethical guidelines. Nevertheless, environmental issues arise as we are aware that both the training and the operation of the BERT model mean an increase in $CO_2$ demand. BERT may not be the latest model, but despite its age, it performed well in our use case and the implementation was easy. Nevertheless, one of the limitations is the model is computationally intensive and requires a lot of resources.

Another limitation was the dependency on Instagram. The goal of the project was to find the best possible way to deliver results to the end user and allow all users to use Instagram more safely. The research team concluded that a Chrome browser extension combined with a REST API is an effective way to ensure results for the average user. However, a major limitation is that it was not possible to embed the results directly into the Instagram app, which is used by most Instagram users. This solution only works on a Chrome browser on a desktop device and major HTML DOM changes on Instagram's website could impact the functionality of our solution.

The problem of false positives is also an important issue for this topic, and the research team concluded that more time should be invested in this matter. Higher precision would further argue for the use of the browser extension.

## Acknowledgements

We would like to thank the Lucerne University of Applied Sciences and Arts and the University of Applied Sciences Technikum Vienna for providing us with the opportunity to conduct our research. This work was funded by both InvestmentExplorer and CryptoExplorer GmbH.

# A  Model Configurations

All code repositories can be found here: https://scamspot.github.io/.

| Configuration Aspect | Details/Settings |
|---|---|
| **Pre-trained Model Name** | bert-base-cased |
| **Tokenizer** | BertTokenizer from bert-base-cased |
| **Epochs** | 10 |
| **Maximum Length** | 512 |
| **Batch Size** | 16 |
| **Data Split** | Train: 80%, Validation: 10%, Test: 10% |
| **Model Class** | ScamClassifier |
| **Optimizer** | AdamW with learning rate 2e-5, correct_bias=False |
| **Scheduler** | Linear schedule without warmup (num_warmup_steps=0) |
| **Loss Function** | CrossEntropyLoss |

Table 3: BERT Model Configurations

| Configuration | Value |
|---|---|
| **Model** | GPT-4 (gpt-4-1106-preview) and GPT-3.5 (gpt-3.5-turbo) |
| **Max Tokens** | 10 |
| **Seed** | 42<br>*Note: As per OpenAI's API documentation, results are not deterministic, even with a seed value and a temperature of 0.* |
| **Temperature** | 0 |
| **F1 Score** | The F1 Score was based on the highest result obtained across three attempts. |
| **System Prompt** | You are a comment moderator at Instagram classifying comments. |
| **User Prompt** | Classify the following Instagram comment as 'spam', 'scam', or 'genuine'. Reply only with the label for this comment: '[comment]' |
| **Mapping** | 'spam', 'scam' = 1<br>'genuine' = 0 |

Table 4: GPT Model Configurations

## B Model & Data Evaluation

| Algorithm | F1 Score | Accuracy | Precision | Recall | ROC AUC Score |
|---|---|---|---|---|---|
| GPT-3 | 0.3699 | 0.5747 | 0.3660 | 0.3739 | 0.5246 |
| GPT-3 | 0.3557 | 0.5689 | 0.3550 | 0.3565 | 0.5160 |
| GPT-3 | 0.3514 | 0.5660 | 0.3506 | 0.3522 | 0.5127 |
| GPT-4 | 0.5911 | 0.7068 | 0.5530 | 0.6348 | 0.6889 |
| GPT-4 | 0.4889 | 0.6328 | 0.4566 | 0.5261 | 0.6062 |
| GPT-4 | 0.3537 | 0.5704 | 0.3553 | 0.3522 | 0.5160 |

Table 5: Model evaluation of GPT-4 (gpt-4-1106-preview) and GPT-3.5 (gpt-3.5-turbo)

## C Existing Instagram Spam Filter

Instagram already hides potential spam messages, which can be displayed again by the user with one click at the end of the comments section. Based on our evaluation, Instagram's existing spam filter has a precision of 98.36%, but only a recall of 11.51%. Posts were selected randomly (Table 6).

**Links & Confusion Matrix**

https://www.instagram.com/p/Cr1eVGbtrZq/
Confusion Matrix: 8,0,9,5

https://www.instagram.com/p/CpskFlcouAm/
Confusion Matrix: 0,0,32,6

https://www.instagram.com/p/CrlTySlPSSL/
Confusion Matrix: 8,1,50,43

https://www.instagram.com/p/CsLCoRIvTNO/
Confusion Matrix: 0,0,32,6

Table 6: Instagram Spam Filter Metrics

## D Example Comments

| |
|---|
| entrepreneurship isn't easy just like protesting when you don't have clue of what's going on that's why i encourage people to passively do something spectacular in case you're seeking for an option on how to make money online get in touch with #hāźeł_mcèwəṅ |
| i have tried several platforms they didn t workout but when i did take the risk to invest $1000 in less than week i got $26 000 from her platform i must confess you are truly the best @trade_with_denise_alvina |
| while waiting for your salary you can earn up to $12 000 in seven working days despite the covid-19 situation you can still make.moremoney without going out @wealthwithmarilynn |
| if you dream of #dogecoin becoming 1$ |

# E User Evaluation

90% of survey participant [N=20] reported that they don't think that Instagram has done enough to combat the current situation with spam and scam on the platform. All users used the browser extension.

| ID | Amount of spam | Has IG done enough | UX Before | Interaction Before | UX After | Interaction After |
|---|---|---|---|---|---|---|
| 1 | A large amount | No | Annoyed, Frustrated | Not likely | Neutral, I still see the comments | Not likely |
| 2 | Almost all of it | No | Worried, It's pretty sad, there are only red comments, nearly all from fraudsters | Unlikely | Interested | Likely |
| 3 | A large amount | No | Annoyed, As always, half of the comment section is spam | Unlikely | Annoyed | Unlikely |
| 4 | A small amount | I'm not sure | Neutral | Not likely | Neutral | Not likely |
| 5 | A large amount | No | Frustrated, Worried | Unlikely | Frustrated | Not likely |
| 6 | Almost all of it | No | Worried | Unlikely | Neutral | Likely |
| 7 | Almost all of it | No | Annoyed | Unlikely | Interested | Likely |
| 8 | A large amount | No | Frustrated, Demotivating | Unlikely | Better, but I want only real comments | Not likely |
| 9 | A large amount | No | Annoyed, Frustrated | Unlikely | Neutral | Not likely |
| 10 | A large amount | No | Worried | Unlikely | Annoyed | Not likely |
| 11 | A large amount | No | Annoyed, Frustrated | Unlikely | Happy, Enthusiastic | Likely |
| 12 | A large amount | No | Annoyed, Frustrated, Worried | Unlikely | Happy, Interested | Very likely |
| 13 | A small amount | I'm not sure | Neutral | Not likely | Interested | Likely |
| 14 | Almost all of it | No | Annoyed | Unlikely | Interested | Likely |
| 15 | A large amount | No | Annoyed, Frustrated | Unlikely | Interested | Likely |
| 16 | Almost all of it | No | Annoyed, Frustrated, No real human connection pssobile, there are only scammers | Unlikely | Happy, Enthusiastic, It's great, I only see real comments | Very likely |
| 17 | Almost all of it | No | Annoyed, Frustrated, I am a page owner myself and I know that this problem has been going on for years, but Instagram does nothing about it. | Unlikely | Happy, Enthusiastic, Love it! | Very likely |
| 18 | Almost all of it | No | Annoyed | Unlikely | Happy, Enthusiastic | Likely |
| 19 | A large amount | No | Annoyed, Frustrated | Unlikely | Interested | Likely |
| 20 | Almost all of it | No | Annoyed, IG doesn't do anything against scams in the crypto space | Unlikely | Enthusiastic, I need this also for Twitter, works great | Very likely |

Table 8: Survey responses of the user evaluation; UX = User Experience

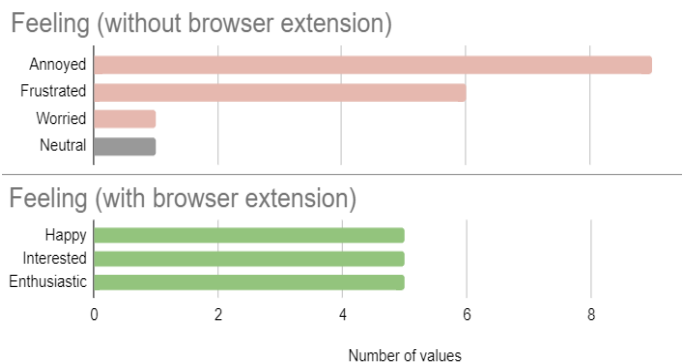

Figure 3: User experience reported without and with the browser extension [N=20]